# An Explainable Anomaly Detection Framework for Monitoring Depression and Anxiety Using Consumer Wearable Devices

Yuezhou Zhang[1], Amos A. Folarin[1,3,4,5,6], Callum Stewart[1], Heet Sankesara[1], Yatharth Ranjan[1], Pauline Conde[1], Akash Roy Choudhury[1], Shaoxiong Sun[2], Zulqarnain Rashid[1], Richard J.B. Dobson[1,3,4,5,6]

[1]Department of Biostatistics & Health Informatics, Institute of Psychiatry, Psychology and Neuroscience, King's College London, UK

[2]Department of Computer Science, University of Sheffield, Sheffield, UK

[3]Institute of Health Informatics, University College London, UK

[4]NIHR Maudsley Biomedical Research Centre, South London and Maudsley NHS Foundation Trust, London, United Kingdom

[5]NIHR Biomedical Research Centre at University College London Hospitals, NHS Foundation Trust, London, United Kingdom

[6]Health Data Research UK London, University College London, London, United Kingdom

Corresponding author: Yuezhou Zhang(yuezhou.zhang@kcl.ac.uk) and Richard Dobson (richard.j.dobson@kcl.ac.uk)

## Abstract

Continuous monitoring of behavior and physiology via wearable devices offers a novel, objective method for the early detection of worsening depression and anxiety. In this study, we present an explainable anomaly detection framework that identifies clinically meaningful increases in symptom severity using consumer-grade wearable data. Leveraging data from 2,023 participants with defined healthy baselines, our LSTM autoencoder model learned normal health patterns of sleep duration, step count, and resting heart rate. Anomalies were flagged when self-reported depression or anxiety scores increased by ≥5 points—a threshold considered clinically significant. The model achieved an adjusted F1-score of 0.80 (precision = 0.73, recall = 0.88) in detecting 393 symptom-worsening episodes across 341 participants, with higher performance observed for episodes involving concurrent depression and anxiety escalation (F1 = 0.84) and for more pronounced symptom changes (≥10-point increases, F1 ~ 0.85). Model interpretability was supported by SHAP-based analysis, which identified resting heart rate as the most influential feature in 71.4% of detected anomalies, followed by physical activity and sleep. Both elevated and unusually low resting heart rates, as well as reduced step counts and shorter sleep durations, were associated with increased anomaly likelihood. We further illustrate individual cases using time-dynamic feature attributions, demonstrating the framework's ability to retrospectively trace the onset and progression of anomalous behavioral or physiological patterns. This approach not only enables users to self-contextualize detected anomalies, but also provides clinicians interpretable insights into the underlying mechanisms of mental disorders. Together, our findings highlight the potential of explainable anomaly detection to enable personalized, scalable, and proactive mental health monitoring in real-world settings.

## Introduction

Depression and anxiety are the most prevalent mental health disorders worldwide [1]. These mental disorders are linked to numerous adverse outcomes, including premature mortality, diminished quality of life, reduced work capacity, disability, and an elevated risk of suicide [2, 3]. Although early interventions can significantly improve outcomes [4], the diagnosis and treatment for depression and anxiety face significant challenges: (1) Current diagnosis methods based on questionnaires or interviews may introduce subjective recall bias [5-7] and fail to capture day-to-day fluctuations in mental status and behaviors [8, 9]; (2) Effective diagnosis relies on skilled mental health professionals, who are in short supply globally, especially in low-income regions [10, 11]; (3) Assessments are often delayed until the conditions worsen to a more severe, difficult-to-treat stage [12], either because initial symptoms are mild and easily overlooked [13] or due to reluctance in seeking help for some specific reasons, e.g. societal stigma [14]. These challenges highlight a crucial need for objective, effective, and scalable methods to detect early changes in the severity of depression and anxiety [15].

Advances in sensors and wearable technology now enable convenient, cost-effective, and accurate monitoring of individual's behaviors (e.g. sleep and physical activity) and physiological signals (e.g. heart rate) [16-18]. Using these tools, several mobile health (mHealth) studies have identified significant associations between depression or anxiety severity and various wearable-derived parameters [19-22]. For instance, higher depression severity has been associated with increased day-to-day variability in sleep duration [23, 24], delayed sleep-wake times [23, 24], reduced physical activity levels [25, 26], increased time spent at home (reduced mobility) [27, 28], disrupted circadian rhythms [29, 30], lower heart rate variability [31, 32], and higher nocturnal heart rates [31, 32]. Many of these patterns have also been observed in individuals with anxiety disorders [21, 33], likely due in part to the high comorbidity between anxiety and depression [34]. Additionally, anxiety severity has been correlated with reduced heart rate variability features [35]. Collectively, these significant associations underscore the potential of using wearable-derived behavioral and physiological features for monitoring changes in depression and anxiety severity.

However, previous mHealth studies using wearable or smartphone data to predict the severity of depression or anxiety have reported widely varying performance [19]. Several studies have also demonstrated the limited capability of cross-sectional predictions based solely on wearable or smartphone data [33, 36, 37]. In addition, Xu et al. replicated previously reported algorithms on their own dataset of 534 participants but observed substantially lower accuracy than originally reported [38]. This inconsistency may stem from multiple factors. First, depression and anxiety are heterogeneous in their presentations, with diverse symptom manifestations across individuals [39, 40]. Second, there is substantial individual difference in behaviors and physiology; for instance, some individuals may have a normal baseline of 10,000 steps per day and 8 hours of sleep, while others may have a baseline of 3,000 steps and 6 hours of sleep. Third, individuals with mild-to-moderate symptoms may not exhibit noticeable behavioral changes every day before assessment, complicating the labeling of training data. For instance, the PHQ-8 depression scale assesses symptoms over the past two weeks with responses ranging from "not at all" to "nearly every day" [41], consequently, individuals reporting moderate symptoms may exhibit “normal” behaviors and physiological patterns on some days prior to the assessment, resulting in noisy labels. Consequently, traditional supervised learning (strictly based on labeled data) on small-medium datasets may produce biased models with poor generalizability for predicting depression and anxiety severity.

To address these challenges, anomaly detection [42] may be one potential solution. Anomaly detection techniques learn the patterns in normal data (health status) and identify deviations (anomalous changes) that significantly differ from expected behaviors, and this approach has been applied in various disease-detection contexts [43, 44]. For example, during the COVID-19 pandemic, many mHealth studies used anomaly detection on consumer wearable data (e.g. from Fitbit or Apple Watch) to enable real-time detection of infection onset by identifying anomalous physiological changes [45-47]. In the context of mental health, a few recent studies have explored anomaly detection for depression and anxiety. Cohen et al. demonstrated that anomalies in smartphone sensor data could predict changes in depression and anxiety scores with acceptable accuracy in two cohorts of 75 individuals [48]. D’Mello et al. analyzed the similarity of behavioral patterns over time in a cohort of 695 college students, finding modest correlations between anomalous changes in routine behaviors and shifts in depression/anxiety questionnaire scores [49]. However, these two studies relied on relatively simple rule-based metrics to define behavioral anomalies, which might not fully capture the complex patterns of real-world behaviors. Vairavan et al. leveraged advanced deep learning technologies to train a personalized anomaly detection model on each individual’s historical wearable-measured activity data for predicting depression relapse [50]. However, this method's reliance on extensive historical data for each participant and bi-monthly clinical visits poses challenges for practical daily monitoring. Furthermore, the "black-box" nature of deep learning-based anomaly detection models highlights the need for explainable approaches that can provide meaningful insights with clinical relevance.

Despite these initial efforts, the application of anomaly detection to mHealth data for monitoring depression and anxiety remains limited. This is likely due to factors such as small-medium sample sizes, short study durations, and a lack of clearly labeled “normal” baseline data in existing studies/datasets. To overcome these limitations, we leveraged a large-scale longitudinal mHealth dataset from a general UK population [51, 52]. Data collection was facilitated by the RADAR-base, an open-source platform developed by our team that integrates passive wearable data with active smartphone questionnaires [53, 54], yielding a substantial amount of data with linked mental health status. The aim of the present study was to develop an explainable deep learning–based anomaly detection framework for predicting anomalous changes in depression and anxiety symptom severity using consumer wearable data.

## Methods

### Study Samples and Settings

We utilized data from Covid Collab, a large-scale observational mHealth study that enrolled 17,667 participants through the Mass Science smartphone app between June 2020 and August 2022 [51]. Participants provided Fitbit wearable data via the RADAR-base platform (Figure 1a) [53], and were also encouraged to regularly complete smartphone-based questionnaires of mental health (depression and anxiety) and COVID-19 symptoms. Detailed information on the study design and procedures is available in the study protocol [51]. Ethical approval for the study was obtained from the PNM Research Ethics Panel at King's College London (LRS-18/19-8662), and all participants provided informed electronic consent through the study app.

### Depression and Anxiety Assessments

Participants self-reported their depression and anxiety symptoms every two weeks via the study app. Depression severity was measured using the 8-item Patient Health Questionnaire (PHQ-8) [41], and anxiety was assessed with the 7-item Generalized Anxiety Disorder scale (GAD-7) [55]. The total scores range from 0 to 24 for PHQ-8 and 0 to 21 for GAD-7, with higher scores indicating more severe symptoms.

### Definition of Normal Period

Since anomalies in behavioral or physiological signals may occur either before or after anomalous changes in depression and anxiety [27, 56], a sufficiently long stable health period is required to capture "normal patterns". We defined a "normal period" for each participant as at least 8 consecutive weeks (4 assessments) during which all PHQ-8 and GAD-7 scores indicated no/minimal symptoms (both below 5 points). Additionally, since COVID-19 can also affect behavior and physiology [52, 57], we required that this 8-week normal period did not overlap with the period from 7 days before to 21 days after any reported COVID-19 infection, as recommended by [47].

**Definition of Anomalous Changes and Period in Depression and Anxiety Severity**

We defined an anomalous change in mental health as a significant increase in a participant's depression or anxiety relative to their normal period. Specifically, an anomalous episode was flagged whenever a participant's PHQ-8 or GAD-7 score exceeded the average score of their normal period by ≥ 5 points (Figure 1b). We chose a 5-point threshold based on clinical evidence that a change of this magnitude in PHQ-8 or GAD-7 is clinically meaningful [58-60]. Anomalous episodes could be categorized into three types based on which score(s) increased: "PHQ-only" for anomalies in depression, "GAD-only" for anomalies in anxiety, and "Both" for anomalies in both depression and anxiety.

For each anomalous episode, we defined a corresponding anomalous period**.** Given that the PHQ-8 and GAD-7 assessments assess symptoms over the past two weeks, the 14 days prior to an anomalous assessment were labeled as anomalous. Furthermore, as prior studies indicating that behavioral or physiological anomalies may precede or follow changes in depressive and anxiety symptoms [27, 56], we empirically extended the window to include the 7 days before and 14 days after this core period. Consequently, each anomalous period was defined as the 21 days preceding and the 14 days following an anomalous assessment.

**Daily Feature Extraction and Data Processing**

From the raw Fitbit recordings, we extracted three key daily features to capture participants' behavioral and physiological patterns:

(1) Sleep Duration: Total time spent asleep each day, calculated as the sum of time in the "light", "deep", and "rapid eye movement" sleep stages recorded by the Fitbit device.
(2) Total Steps: The total number of steps recorded by the Fitbit each day, serving as a proxy for overall daily physical activity.
(3) Resting Heart Rate: Daily resting heart rate was computed using an established algorithm [45-47]. Specifically, we identified all periods of at least 12 consecutive minutes with zero step counts (indicative of resting period) and computed the daily resting heart rate as the average heart rate during these periods.

We excluded days with more than 20% missing data for either step count or heart rate from feature calculations [29, 33]. Missing values in the daily features were imputed using linear interpolation. Finally, to account for individual differences in baseline levels and variability, we applied z-score normalization to each of the three features on a per-participant basis.

**Anomaly Detection Model**

The Long Short Term Memory Network-based autoencoder (LSTM-AE) is a widely used approach for anomaly detection in time-series data, designed to learn patterns inherent in normal data [61, 62]. By leveraging LSTM networks, the LSTM-AE captures temporal dependencies and interrelationships between features across time steps. It consists of two main components: an encoder that compresses input time-series data into a fixed-length latent vector, and a decoder that reconstructs the original sequences. For normal data, the LSTM-AE can accurately reconstruct the input, resulting in low reconstruction error. In contrast, for anomalous data that deviate significantly from the learned patterns, the reconstruction error is substantially higher.

In this study, we segmented the time-series data—comprising three daily features (Sleep Duration, Total Steps, and Resting Heart Rate)—using a 7-day sliding window with a 1-day moving step to generate input sequences for the LSTM-AE model. The model was trained exclusively on “normal period” data (Figure 1c). To ensure data quality, only normal data with less than 20% missingness in daily features were included in the training process. The normal data were split into 80% training and 20% validation sets. To prevent overfitting, we applied an early stopping strategy, halting training if the reconstruction error on the validation set did not decrease for 10 consecutive epochs. We used the same model architecture and parameters as reported in [61, 62].

Previous studies have often used the maximum reconstruction error on the validation set as a threshold for detecting anomalies [47, 63]. However, in the context of mHealth, behaviors and physiology may be influenced by factors such as illness, workload, or travel even during the "normal period," which were not recorded in this study. To account for this, we explored various percentile thresholds of the reconstruction error distribution on the validation set, specifically the 90th to 100th percentiles.

**Evaluation Metrics**

Anomalous changes in depression and anxiety may only affect behaviors and physiology on certain days within the anomalous period. For instance, the PHQ-8 assesses symptoms over the past two weeks with responses of "not at all", "several days", "more than half the days", and "nearly every day", indicating that a participant with moderate symptoms may still exhibit relatively "normal" behaviors and physiology on some days during the anomalous period. To account for this, we adopted the adjusted F-score proposed by Xu et al. for evaluating anomaly detection in an event-based context [64]. This evaluation method has been also applied in other studies on anomaly detection for health events [65, 66]. Under this approach, an anomalous episode is considered successfully detected if the model flags at least one input segment (7-day window) as anomalous within that episode. We report the adjusted F-score along with its associated precision and recall.

**Model Interpretability**

The SHapley Additive exPlanations (SHAP) method is a widely used approach for model interpretation [67] and is also leveraged to explain anomaly detection in autoencoders [68].

The contribution of each feature to a prediction outcome is represented by its SHAP values, with the magnitude reflecting the feature's importance [67]. In this study, the outcome of the LSTM-AE model is the reconstruction error. Since a higher reconstruction error suggests a greater likelihood of anomalies, the SHAP values for each variable can indicate their relationship with anomalous changes.

To explore the variability of feature contributions across different individuals/types of categories, we ranked features based on their contribution in each anomaly episode. We summarize feature ranks across all episodes and compared them between different anomaly categories using chi-square tests [69]. Additionally, we visualized SHAP values over time for each feature to explore the origins of anomalies and the causes of false alarms.

## Results

According to our criteria, a total of 314,960 days from 2,023 participants were classified as the normal period, defined as at least 8 consecutive weeks with PHQ-8 and GAD-7 scores below 5 and no reported COVID-19 infections. A total of 393 anomalous episodes from 341 participants were identified, where PHQ-8 or GAD-7 scores increased by ≥5 points compared to their normal period. Among these, 100 episodes involved anomalies in both PHQ-8 and GAD-7 (denoted as BOTH), 148 in PHQ-8 only (PHQ-only), and 145 in GAD-7 only (GAD-only). For the magnitude of change, 214 episodes showed a 5–9 point increase in PHQ-8, while 34 episodes increased by ≥10 points. Similarly, 209 episodes had a 5–9 point increase in GAD-7, and 36 episodes increased by ≥10 points. Table 1 provides a summary of participant demographics and the distribution of anomaly categories.

Figure 2a visualizes the average temporal changes (normalized by participants) in sleep, step count, and resting heart rate across all anomaly episodes. Notably, daily total step count shows a considerable decline, and resting heart rate increases during the anomalous period, while changes in sleep duration are relatively small.

The LSTM-AE model, trained on normal period data from 2,023 participants, achieved its highest performance with an adjusted F-score of 0.7953, a precision of 0.7277, and a recall of 0.8768 when using the 95th percentile of validation loss as the detection threshold across all anomaly episodes. Performance was better for BOTH anomalies (F = 0.8375; Precision = 0.7660; Recall = 0.9237) compared to PHQ-only (F = 0.7679; Precision = 0.6999; Recall = 0.8549) and GAD-only (F = 0.7842; Precision = 0.7224; Recall = 0.8576). Additionally, the model performance was better for detecting anomalies with ≥10-point increases (F = 0.8527 for PHQ-8 and F = 0.8515 for GAD-7) compared to those with 5–9 point increases (F = 0.7912 for PHQ-8 and F = 0.8006 for GAD-7). These performance metrics are illustrated in Figures 2b and Supplementary Figure 1.

We calculated SHAP values to evaluate the contribution and importance of different wearable-derived features to the model outcome (reconstruction error) for all anomaly episodes. Overall, resting heart rate had the highest feature importance, followed by total steps and sleep duration (Figure 3a and 3b). SHAP dependence plots revealed a U-shaped relationship between reconstruction error and resting heart rate, where both excessively high and low values were associated with higher reconstruction error (Figure 3c). Additionally, reconstruction error showed a negative correlation with total steps (Figure 3d) and sleep duration (Figure 3e).

We also analyzed the contribution rank of each feature for individual anomalous episodes, finding that Resting Heart Rate ranked first in 71.4% of episodes, Total Steps in 20.3%, and Sleep Duration in 8.3% (Figure 3f-3h). Feature rank distribution varied across anomaly categories: for Sleep Duration, the contribution was significantly higher in GAD-only anomalies (Rank 2: 22.8%) compared to BOTH (Rank 2: 6.1%) and PHQ-only anomalies (Rank 2: 14.6%) ($\chi^2$ test: p = 0.006) (Figure 3g). For Total Steps, the contribution was considerably higher in BOTH anomalies (Rank 2: 73.7%) compared to PHQ-only (Rank 2: 61.6%) and GAD-only (Rank 2: 56.4%) (p = 0.058) (Figure 3h). In contrast, the rank distributions of Resting Heart Rate remained similar across anomaly categories (p = 0.85) (Figure 3f).

To further interpret anomaly origins and identify causes of false alarms, we analyzed SHAP values over time for each feature. Figure 4 presents four examples of anomaly detections, illustrating SHAP values over time. Figure 4a shows a PHQ-8 anomaly accompanied by clear changes in all three features: reduced and irregular sleep, decreased step count, and increased resting heart rate. Figure 4b shows another PHQ-8 anomaly, mainly characterized by a notable decrease in step count and a moderate increase in resting heart rate. Figure 4c illustrates a GAD-7 anomaly, where decreased and irregular sleep preceded a subsequent rise in resting heart rate. Finally, Figure 4d demonstrates a successful detection alongside a false alarm, where the false positive was primarily driven by temporary fluctuations in sleep patterns over a few days.

## Discussion

This study introduces a novel, explainable time-series anomaly detection framework for identifying anomalous changes in depression and anxiety using daily wearable-derived features. By exclusively training on "normal data", this approach mitigates biases associated with the heterogeneity in mental disorder symptoms and the potential inaccuracy of mental health labels. The framework not only captures complex temporal dynamics but also enhances interpretability, allowing clinicians and data scientists to trace the origins of anomalies and gain insights into the diverse manifestations of mental health changes. By reinforcing known behavioral and physiological indicators, this study advances the understanding of depression and anxiety detection while providing a more nuanced perspective on how these indicators co-occur in real-world settings.

The behavioral and physiological anomalies identified by our framework align with previous clinical and mobile health studies, supporting its validity and ability to capture clinically relevant changes rather than arbitrary outliers. For instance, consistent with our findings, elevated resting heat rates have been widely documented in individuals with depression and anxiety [70-72], potentially reflecting chronic stress [73], autonomic nervous system dysregulation [74], and heightened sympathetic nervous system activation [75]. This increased sympathetic drive is associated with physiological hyperarousal symptoms, commonly experienced by individuals with anxiety disorders [76, 77], and has also been linked to inflammation and cardiovascular risks commonly observed in individuals with depression [75, 78]. Furthermore, we also found that excessively low resting heart rates were associated with mental health anomalies, with similar findings have been reported in other studies [31]. While further validation and exploration of the underlying mechanisms of this nonlinear relationship are needed, it may help explain inconsistencies in prior research findings [77].

Likewise, we found that reduced physical activity (approximated by step count) and shorter sleep duration are associated with anomalous changes in depression and anxiety, consistent with previous findings. Reduced physical activity has been widely associated with mental disorders [79, 80]. Previous studies have also linked physical inactivity to disruptions in dopamine release and endorphin production, as well as increased systemic inflammation and hypothalamic-pituitary-adrenal (HPA) axis dysregulation, all of which are associated with mental disorders [81]. The negative association between physical activity and symptom severity has been reported in many mHealth studies [25, 82, 83]. Sleep disturbances, including insufficient sleep and insomnia, are both symptoms and risk factors for depression and anxiety, reflecting a bidirectional relationship [84]. Poor sleep quality has been associated with emotional dysregulation, increased stress sensitivity, impaired cognitive function, and heightened sympathetic nervous system activation [85, 86], which are strongly linked to depression and anxiety. The association between shorter sleep and higher severity was also reported in previous mHealth studies [23, 87].

Beyond reinforcing existing associations, our explainable anomaly detection framework provides time-dynamic interpretations across individuals and anomaly types. Comparing feature ranks measured by SHAP method across all anomaly episodes, we identified the resting heart rate, reflecting physiological arousal, as a more universal digital biomarker of mental health changes. This may be because resting physiological signals are less influenced by daily life variations (e.g., travel, holidays, workload shifts) compared to step count and sleep. In addition, our analysis notably suggested potential differences between depression-related and anxiety-related anomalies. Sleep feature ranks higher in anxiety-related episodes compared to those with only depression or both depression and anxiety anomalies. While this finding requires further validation, our framework provides a potential approach to distinguishing specific indicators of depression and anxiety. Additionally, we observed that detection performance was higher when both depression and anxiety worsened simultaneously, possibly because co-occurring symptoms manifest more prominently in behavioral and physiological changes. Furthermore, detection performance was higher for more severe anomalies (≥10-point increases) compared to moderate ones (5–9 points), though the limited number of severe cases suggests the need for further validation in future research.

False alarm is always a challenge in anomaly detection in mHealth studies. Behavioral and physiological features can be influenced by various factors, including lifestyle events such as travel, vacations, workload adjustments, illness, excessive exercise, alcohol consumption, and activities (e.g., parties) [46]. These events may cause deviations in behavior or physiology that are unrelated to mental health changes. Our explainable framework provides clinicians, data scientists, and users with a retrospective and visual tool to analyze the source of each alarm. For example, Figure 4d illustrates that fluctuations in sleep duration were the primary cause of a false alarm. Users can easily self-contextualize the detected anomalies, thereby improving usability [46]. Furthermore, our time-dynamic interpretations offer a way to explore the sequence of behavioral and physiological anomalous changes. For instance, in Figure 4c, sleep disturbances preceded heart rate anomalies, suggesting a potential temporal pattern in mental health changes. While further research is needed, this approach provides a framework for investigating the underlying mechanisms of mental disorders.

This study has several limitations. First, although our model was trained on “normal data” from relatively large samples (over 2,000 participants), most were from the United Kingdom, potentially limiting the generalizability of the learned patterns. Future studies should validate the model on more diverse datasets and include participants from a broader range of backgrounds. Second, as the data was obtained from a self-enrolled observational study, participants could leave at any time, resulting in varying data collection durations. Due to insufficient data for individual-level training and fine-tuning, we applied feature normalization on each participant to reduce the impact of individual differences. In future studies with longer baseline periods, individual-level fine-tuning could help capture personal behavioral and physiological patterns more accurately. Third, in the absence of similar existing studies, some definitions for anomaly detection (such as the magnitude of change, anomaly duration, and detection thresholds) were determined empirically and require further investigation. Fourth, to demonstrate the feasibility of the framework, we adopted a relatively simple time-series anomaly detection model and focused on daily-level features for clinical interpretability. Future work could explore more advanced models and higher-resolution data, such as minute-level or hourly features, to capture more detailed patterns.

In conclusion, the proposed anomaly detection framework demonstrated robust performance in identifying clinically meaningful increases in depression and anxiety, while providing interpretable insights into behavioral and physiological changes. These findings highlight the feasibility of scalable, low-burden monitoring using consumer wearables to support early detection, personalized care, and timely intervention in mental health.

**Data availability**

De-identified participant data are available for academic research purposes upon request to the corresponding author and the signing of a data access agreement.

**Code availability**

The complete code used for the analysis can be made available through reasonable requests. Please email the corresponding author for details.

**Author contribution**

Y.Z., A.A.F., and R.J.B.D. designed this study. Y.Z. conducted the data analyses and drafted the manuscript. C.S., Y.R., P.C., A.A.F., Z.R., and R.J.B.D. contributed to the data collection. H.S. and S.S. provided input on data analyses. A.A.F., C.S., H.S., Y.R., P.C., A.R.C., S.S., Z.R., and R.J.B.D. contributed to the interpretation of findings and manuscript review. All authors have read and approved the manuscript.

**Competing interests**

Amos A. Folarin reports holding shares in Google, the parent company of Fitbit, which produces the wearable devices utilized in the Covid-Collab study to collect data. Fitbit advertised the Covid-Collab study in the UK Fitbit app. Neither Google nor Fitbit provided funding or devices for this study. All other authors declare no competing interests.

**Acknowledgements**

This study represents independent research partly funded by the National Institute for Health and Care Research (NIHR) Maudsley Biomedical Research Centre (IS-BRC-1215-20018 and NIHR203318) at South London and Maudsley NHS Foundation Trust, Medical Research Council, UK Research and Innovation, and King’s College London. The views expressed in this paper are those of the authors and not necessarily those of the NIHR or the UK Department of Health and Social Care. Richard J.B. Dobson is supported by the following: (1) National Institute for Health and Care Research (NIHR) Biomedical Research Centre (BRC) at South London and Maudsley National Health Service (NHS) Foundation Trust and King's College

London; (2) Health Data Research UK, which is funded by the UK Medical Research Council (MRC), Engineering and Physical Sciences Research Council, Economic and Social Research Council, Department of Health and Social Care (England), Chief Scientist Office of the Scottish Government Health and Social Care Directorates, Health and Social Care Research and Development Division (Welsh Government), Public Health Agency (Northern Ireland), British Heart Foundation, and Wellcome Trust; (3) the BigData@Heart Consortium, funded by the Innovative Medicines Initiative 2 Joint Undertaking (which receives support from the EU's Horizon 2020 research and innovation programme and European Federation of Pharmaceutical Industries and Associations [EFPIA], partnering with 20 academic and industry partners and European Society of Cardiology); (4) the NIHR University College London Hospitals BRC; (5) the NIHR BRC at South London and Maudsley (related to attendance at the American Medical Informatics Association) NHS Foundation Trust and King's College London; (6) the UK Research and Innovation (UKRI) London Medical Imaging & Artificial Intelligence Centre for Value Based Healthcare (AI4VBH); and (7) the NIHR Applied Research Collaboration (ARC) South London at King's College Hospital NHS Foundation Trust.

**Table 1. A summary of demographics of included participants and types of flagged anomaly episodes.**

| Characteristics | Value |
|---|---|
| **Participants with normal periods (for training)** | |
| N | 2023 |
| Age, median (IQR) | 55.0 (46.0-63.0) |
| Female, n (%) | 1205 (59.6) |
| BMI, median (IQR) | 25.6 (23.0-28.7) |
| **Participants with anomaly episodes** | |
| N | 341 |
| Age, median (IQR) | 53.5 (44.8-62.0) |
| Female, n (%) | 255 (74.8) |
| BMI, median (IQR) | 25.0 (22.7-28.5) |
| **Anomaly episodes, N** | |
| All types | 393 |
| BOTH (anomaly in both PHQ-8 and GAD-7) | 100 |
| PHQ-only (anomaly in PHQ-8 only) | 148 |
| GAD-only (anomaly in GAD-7 only) | 145 |
| 5-9-point increase in PHQ-8 | 214 |
| ≥ 10-point increase in PHQ-8 | 34 |
| 5-9-point increase in GAD-7 | 209 |
| ≥ 10-point increase in GAD-7 | 36 |

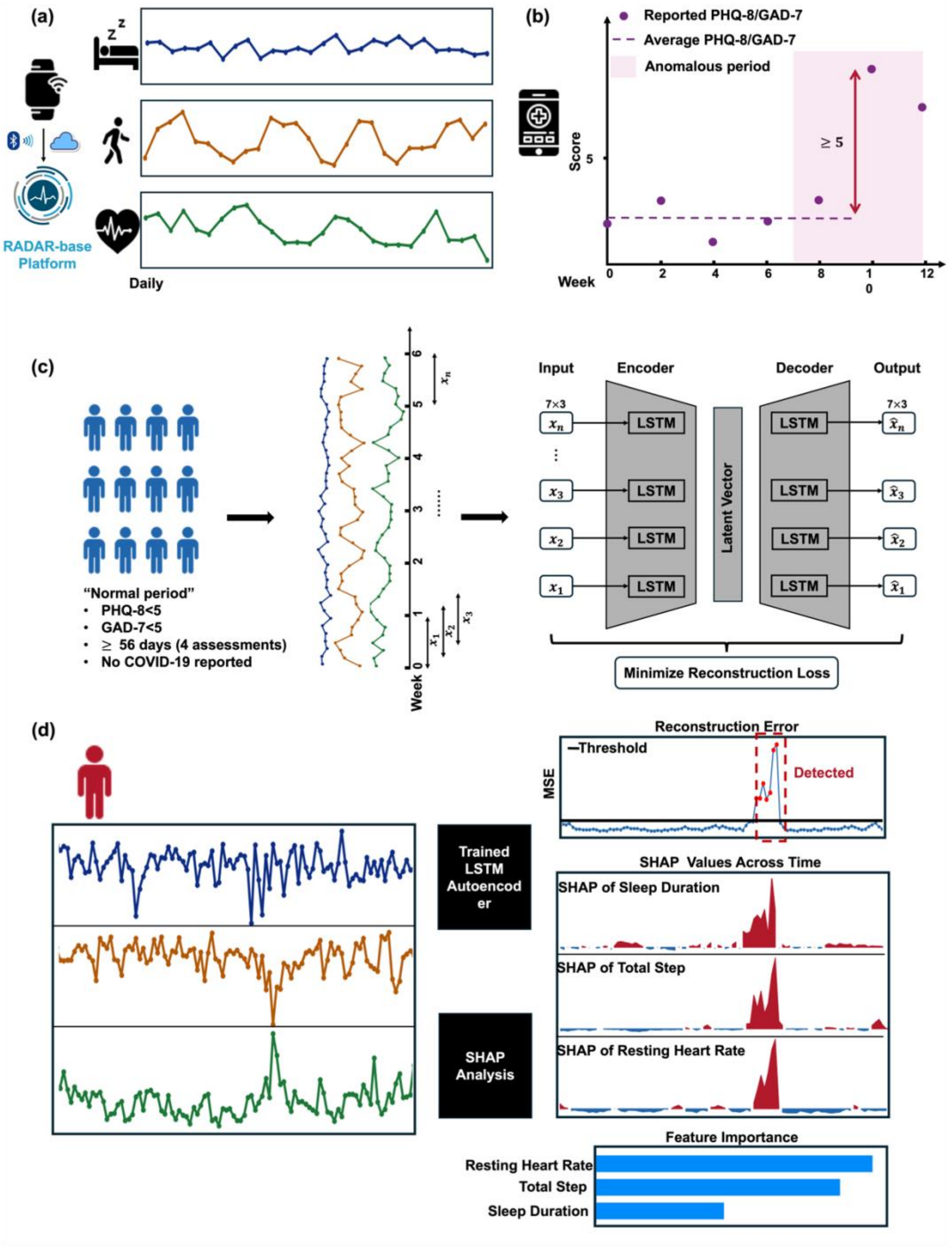

**Figure 1. Overview of the study design and the explainable anomaly detection framework.** (a) Schematic of the Covid Collab mHealth study data collection. (b) Definition of anomalous episodes based on a ≥5-point increase in PHQ-8 or GAD-7 from the participant's normal baseline. (c) LSTM autoencoder model training pipeline using only normal-period data. (d) An anomaly is detected when a new sequence's reconstruction error exceeds a threshold and time-dynamic interpretations for tracing the origins of the detected anomaly.

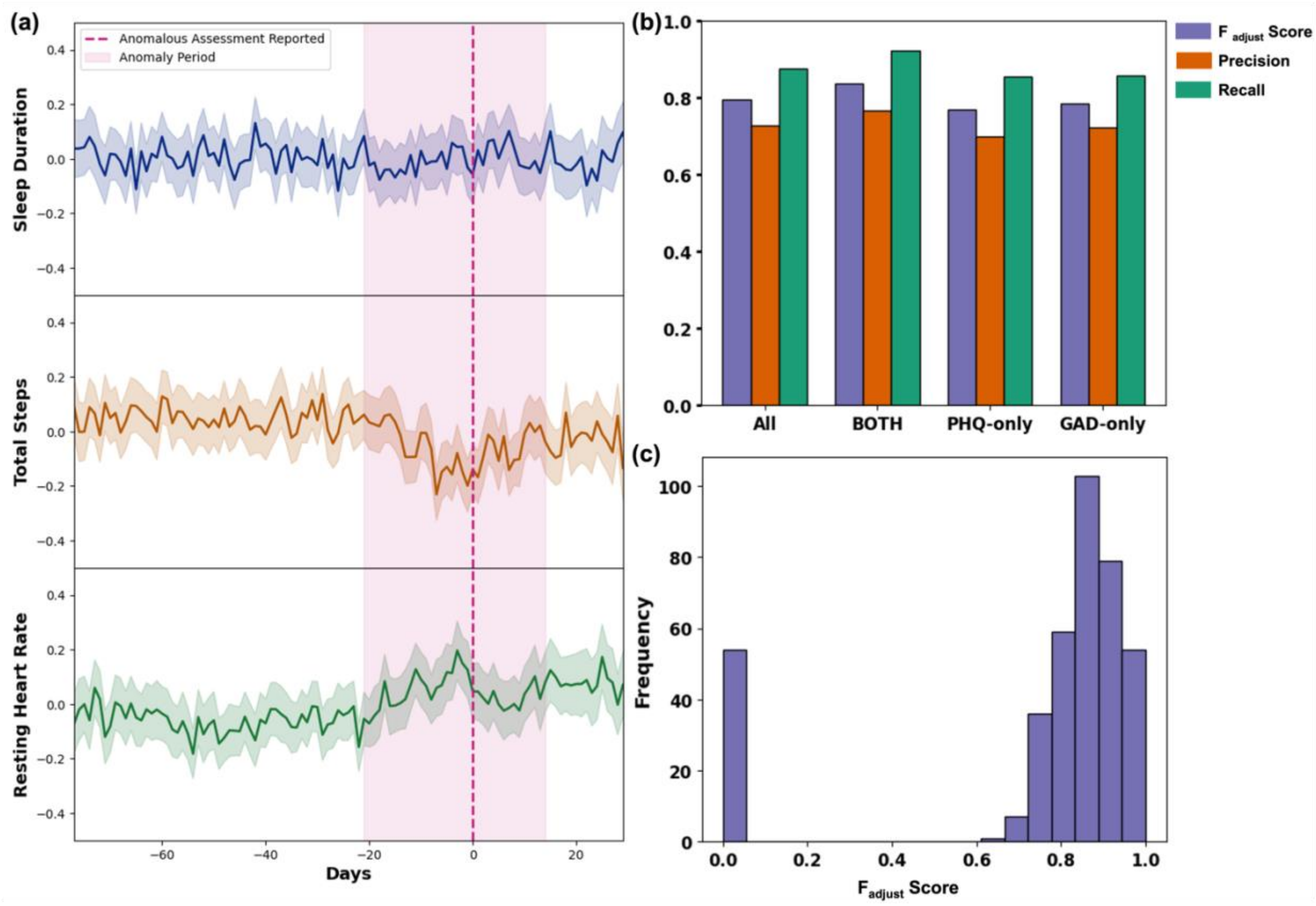

**Figure 2. Behavioral and physiological changes during anomalous episodes and model performance.** (a) Time series of wearable-derived daily features—sleep duration (top), total steps (middle), and resting heart rate (bottom)—centered around the anomalous assessment (dashed pink line). (b) Performance metrics of the anomaly detection model across different anomaly types. (c) Distribution of adjusted F1-scores across all 393 anomalous episodes. A total of 54 episodes had an adjusted F1-score of 0, indicating they were not detected, yielding an overall detection success rate of 86.3%.

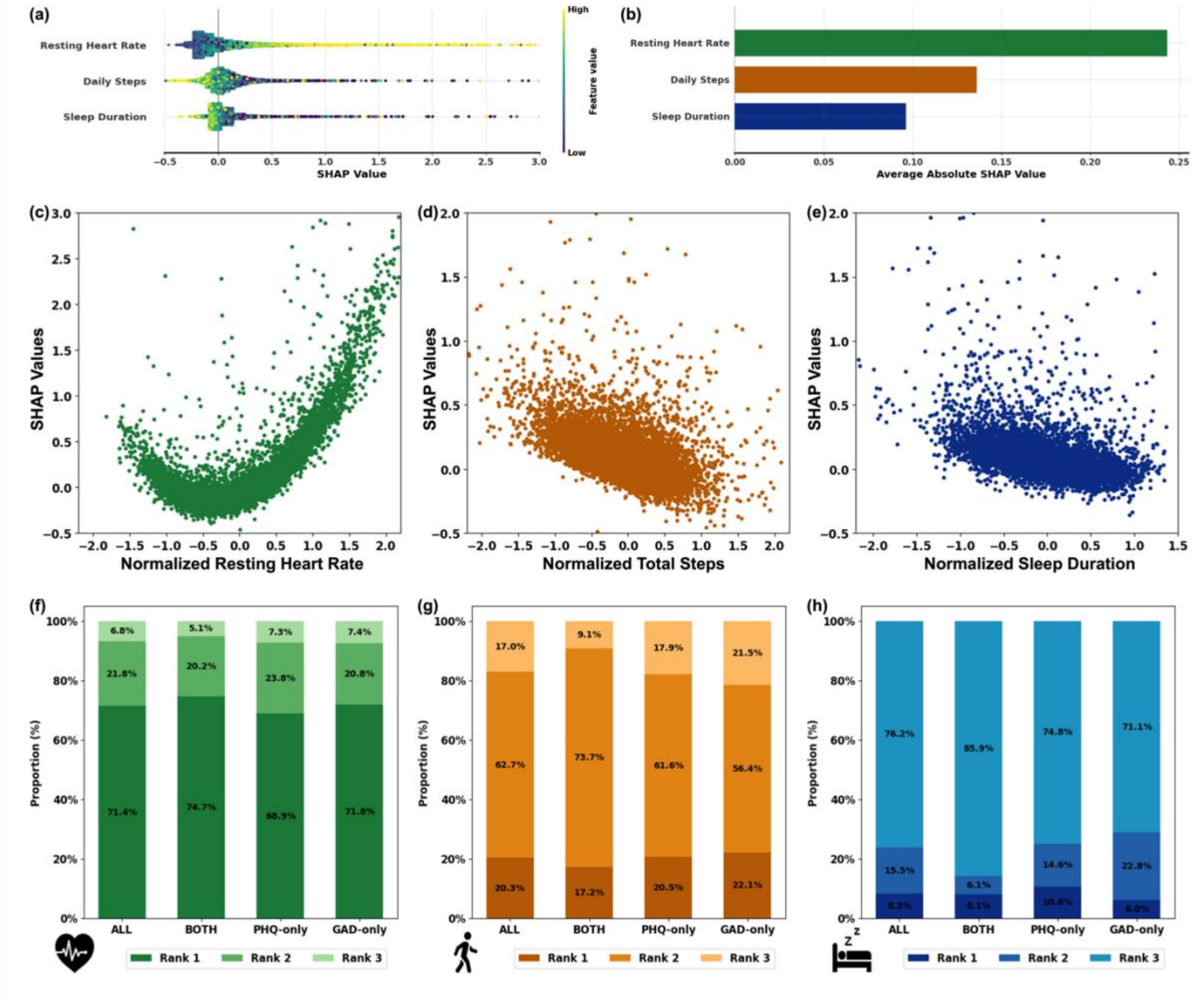

**Figure 3. SHAP feature importance and contribution rankings across anomaly episodes.** (a, b) Overall feature importance derived from SHAP values, indicating that resting heart rate is the strongest contributor to anomaly detection, followed by step count and then sleep duration. (c, d, e) SHAP dependence plots for resting heart rate, step count and sleep duration, respectively, illustrating their relationships with reconstruction error. Higher reconstruction errors suggest greater anomaly likelihood. (f, g, h) Distribution of feature ranks across all anomaly episodes, showing the relative importance of each wearable-derived feature in detecting anomalies.

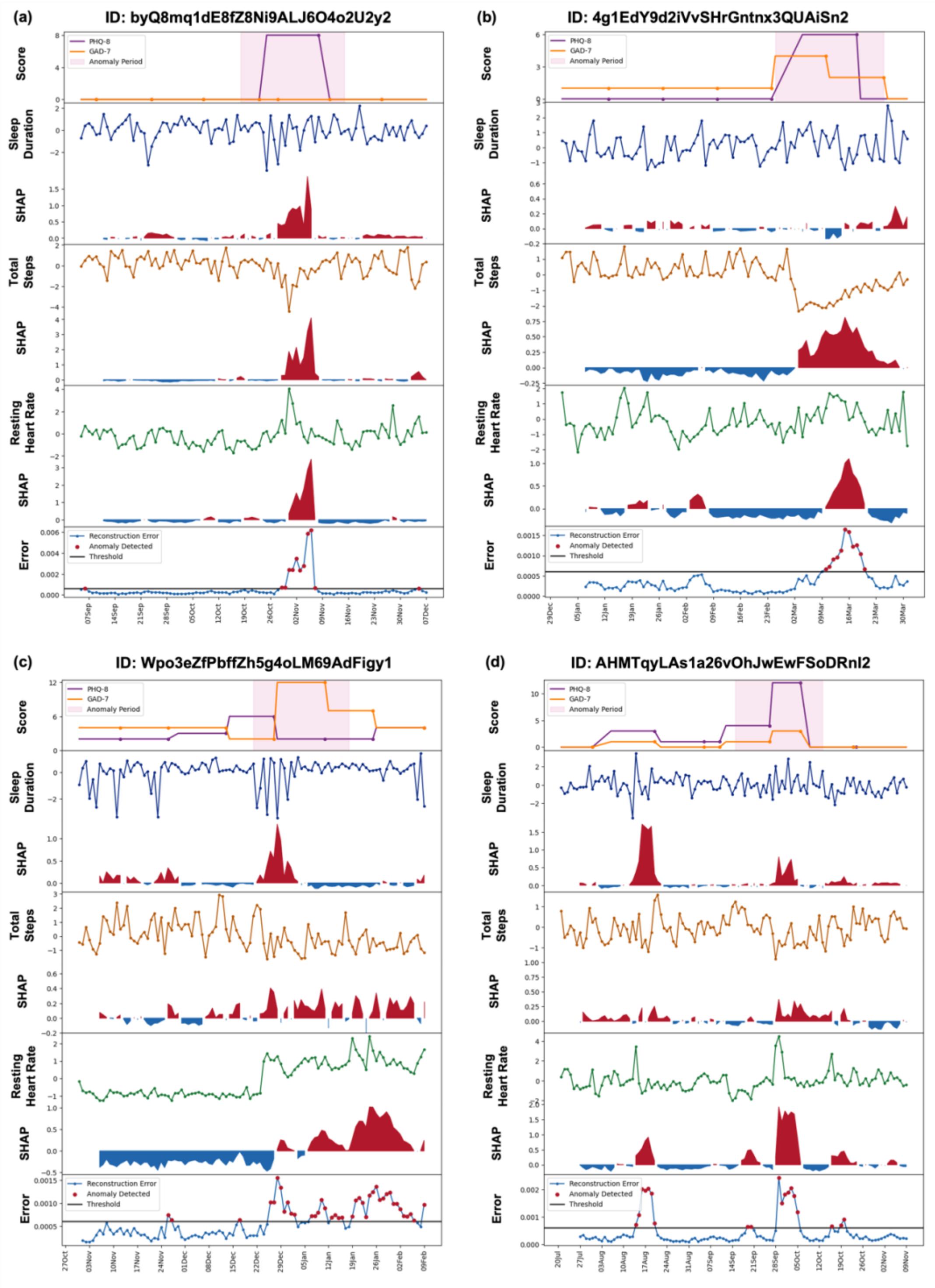

**Figure 4. Examples of detected anomalies with time-dynamic SHAP explanations.** (a) A depression-related anomaly characterized by clear changes across all three features. (b) Another depression-related anomaly primarily driven by a sharp decline in step count and a moderate increase in resting heart rate. (c) An anxiety-related anomaly in which sleep disturbances preceded a delayed rise in resting heart rate. (d) A successful detection alongside a false alarm; the false alarm was largely attributable to temporary sleep fluctuations.